\documentclass[english]{article}
\usepackage{geometry}

\usepackage[T1]{fontenc}
\usepackage{subcaption}
\usepackage[latin9]{inputenc}
\usepackage{bm}
\usepackage{amsmath,mathtools}
\usepackage{amssymb}
\usepackage[unicode=true,
 bookmarks=false,
 breaklinks=false,pdfborder={0 0 1},colorlinks=false]
 {hyperref}
\hypersetup{
 colorlinks,citecolor=blue,filecolor=blue,linkcolor=blue,urlcolor=blue}

\usepackage{enumitem}

\makeatletter
\usepackage{amsthm}
\usepackage{comment}
\usepackage{natbib}
\usepackage{booktabs}

\usepackage{graphicx}
\usepackage{cleveref}
\usepackage[linesnumbered,ruled,vlined]{algorithm2e}

\SetCommentSty{mycommfont}
\usepackage{algorithmic}

\usepackage{float}
\usepackage{multirow}
 
\usepackage{dsfont}
\usepackage{tcolorbox}

\usepackage{color}
\definecolor{yxc}{RGB}{255,0,0}
\definecolor{yjc}{RGB}{125,0,0}
\definecolor{ytw}{RGB}{255,69,0}
\definecolor{gen}{RGB}{0,0,200}
\definecolor{xxy}{RGB}{125,125,0}

\allowdisplaybreaks

\definecolor{yanxi}{RGB}{0,200,100}

\theoremstyle{plain}

\newtheorem{theo}{Theorem}[section]

\newtheorem{lem}{Lemma}[section]
\newtheorem{prop}{Proposition}[section]
\newtheorem{cor}{Corollary}[section]

\theoremstyle{definition} 

\newtheorem{nota}{Notation}[section]
\newtheorem{de}{Definition}[section]
\newtheorem{exa}{Example}[section]
\newtheorem{as}{Assumption}[section]
\newtheorem{alg}{Algorithm}[section]

\newcommand{\btheo}{\begin{theo}}
\newcommand{\bde}{\begin{de}}
\newcommand{\ble}{\begin{lem}}
\newcommand{\bpr}{\begin{prop}}
\newcommand{\bno}{\begin{nota}}
\newcommand{\bex}{\begin{exa}}
\newcommand{\bcor}{\begin{cor}}
\newcommand{\spro}{\begin{proof}}
\newcommand{\bas}{\begin{as}}
\newcommand{\balg}{\begin{alg}}

\newcommand{\etheo}{\end{theo}}
\newcommand{\ede}{\end{de}}
\newcommand{\ele}{\end{lem}}
\newcommand{\epr}{\end{prop}}
\newcommand{\eno}{\end{nota}}
\newcommand{\eex}{\end{exa}}
\newcommand{\ecor}{\end{cor}}
\newcommand{\fpro}{\end{proof}}
\newcommand{\eas}{\end{as}}
\newcommand{\ealg}{\end{alg}}

\theoremstyle{plain}

\newtheorem{theos}{Theorem}
\newtheorem{props}{Proposition}
\newtheorem{lems}{Lemma}
\newtheorem{cors}{Corollary}

\theoremstyle{definition}
\newtheorem{exas}{Example}
\newtheorem{algs}{Algorithm}
\newtheorem{asss}{Assumption}
\newtheorem{defns}{Definition}

\newcommand{\btheos}{\begin{theos}}
\newcommand{\etheos}{\end{theos}}
\newcommand{\bprops}{\begin{props}}
\newcommand{\eprops}{\end{props}}
\newcommand{\bdes}{\begin{defns}}
\newcommand{\edes}{\end{defns}}
\newcommand{\blems}{\begin{lems}}
\newcommand{\elems}{\end{lems}}
\newcommand{\bcors}{\begin{cors}}
\newcommand{\ecors}{\end{cors}}
\newcommand{\bexs}{\begin{exas}}
\newcommand{\eexs}{\end{exas}}
\newcommand{\balgs}{\begin{algs}}
\newcommand{\ealgs}{\end{algs}}
\newcommand{\bass}{\begin{asss}}
\newcommand{\eass}{\end{asss}}

\title{Leveraging Multimodal Diffusion Models
to Accelerate Imaging with Side Information}

\author{
  \begin{tabular}{ccccc}
    \shortstack{Timofey Efimov\thanks{Department of Electrical and Computer Engineering, Carnegie Mellon University, Pittsburgh, PA 15213, USA; Emails: \texttt{\{tefimov, harryd,yuejiec\}@andrew.cmu.edu}.}\\
    CMU}
 &\qquad &
   \shortstack{Harry Dong\footnotemark[1]\\
    CMU} 
 &\qquad &
   \shortstack{Megna Shah\thanks{Materials and Manufacturing Directorate, Air Force Research Laboratory, Wright-Patterson AFB, OH 45433, USA; Emails: \texttt{\{megna.shah.1,jeff.simmons.3, sean.donegan\}@afrl.af.mil}.}\\
    AFRL}
    \\
    \\
    \shortstack{Jeff Simmons\footnotemark[2]\\
    AFRL}
 &\qquad &
   \shortstack{Sean Donegan\footnotemark[2]\\
    AFRL}
 &\qquad &
   \shortstack{Yuejie Chi\footnotemark[1]\\
    CMU}
\end{tabular}
}

\date{\today}

\makeatother

\begin{document}

\theoremstyle{plain} \newtheorem{lemma}{\textbf{Lemma}}\newtheorem{proposition}{\textbf{Proposition}}\newtheorem{theorem}{\textbf{Theorem}}

\theoremstyle{assumption}\newtheorem{assumption}{\textbf{Assumption}}
\theoremstyle{remark}\newtheorem{remark}{\textbf{Remark}}
\theoremstyle{definition}\newtheorem{definition}{\textbf{Definition}}

\maketitle

\begin{abstract}	
Diffusion models have found phenomenal success as expressive priors for solving inverse problems, but their extension beyond natural images to more structured scientific domains remains limited. Motivated by applications in materials science, we aim to reduce the number of measurements required from an expensive imaging modality of interest, by leveraging side information from an auxiliary modality that is much cheaper to obtain. To deal with the non-differentiable and black-box nature of the forward model, we propose a framework to train a multimodal diffusion model over the joint modalities, turning inverse problems with black-box forward models into simple linear inpainting problems. Numerically, we demonstrate the feasibility of training diffusion models over materials imagery data, and show that our approach achieves superior image 
 reconstruction by leveraging the available side information, requiring significantly less amount of data  from the expensive microscopy modality.
\end{abstract}




\section{Introduction}
\label{sec:intro}

The rise of diffusion models\citep{ddpm, song2020score, song2020denoising, sohl2015deep} has revolutionized the field of generative modeling, achieving the state-of-the-art performance for many tasks such as text-to-image generation\citep{podell2023sdxl,zhang2023text}, video generation\citep{ho2022video}, and image editing\citep{kawar2023imagic}. Despite holding a great potential, leveraging diffusion models for inverse problems with scientific or highly structured data is yet to be thoroughly explored. 
Specifically,
many signal processing and computational imaging applications face the problem of inferring an unknown image \( x^\star \in \mathbb{R}^d \) from a small number of its partial observations, which is inadequate to specify $x^{\star}$ without invoking additional constraints. 
The wide success of diffusion models for unconditional generation has motivated their use in regularizing such ill-posed image reconstruction tasks by serving as an expressive prior distribution for the image of interest. Notable algorithms for inverse problems with diffusion priors include DPS\citep{chung2022diffusion}, RED-Diff\citep{mardani2023variational}, DDRM\citep{kawar2022denoising}, DPnP\citep{xu2024provably}, RePaint\citep{lugmayr2022repaint}, to name only a few. 

This paper is motivated by applications where side information---available in the form of auxiliary modality---is available to help accelerate the imaging process. Specifically, let 
\( y \in \mathbb{R}^m \) be measurements from a forward model \( f : \mathbb{R}^d \rightarrow \mathbb{R}^m \), possibly corrupted by measurement noise \( \xi \in \mathbb{R}^m \):
\begin{align} \label{eq:forward}
     y = f(x^\star) + \xi. 
\end{align}
While direct measurements of $x^{\star}$ could be difficult or expensive to collect, the side information $y$ is a cheaper and more readily available modality containing useful information regarding $x^{\star}$. For example, in materials science, while the collection of electron backscatter diffraction (EBSD) images of a material is time-consuming, polarized light (PL) images are much cheaper and easier to collect, while containing useful information regarding the EBSD images \citep{jin2018correlation,jin2020c}. Similar examples can be found in other domains such as medical imaging \citep{islam2023improving,lyu2209conversion}, where it is helpful to harness low-cost imaging modalities to reduce the acquisition burden of the more costly, but related, one.

However, one major and prevalent challenge of fusing the multimodal data in image reconstruction is that the forward
model \eqref{eq:forward} linking the modalities are often non-differentiable and accessible only through a black-box nature for zeroth-order evaluation. 
Unfortunately, existing image reconstruction methods with diffusion models usually assume access to a differentiable forward model, making them not directly applicable. To combat these challenges, we propose to train multimodal diffusion models to reconstruct modalities with arbitrary black-box forward models, essentially turning these nonlinear inverse problems into linear inpainting problems. We make the following contributions.
\begin{itemize}
    \item We reformulate a nonlinear inverse problem with black-box forward models into a linear inpainting problem by training a multimodal diffusion model to capture the joint distribution of different modalities.
    \item We validate our approach in a materials science application for EBSD image reconstruction with PL images as side information, and demonstrate  its superior performance compared to a unimodal model trained with similar resources, highlighting benefits of additional modalities. 
    \item Our work extends the reach of diffusion models to materials imagery data, highlighting the potential of generative models for dealing with highly structured scientific data.
\end{itemize}




The rest of this paper is organized as follows.
In Section~\ref{sec:background}, we cover relevant background in diffusion models and inverse problems using diffusion models as a prior. Then, we introduce our algorithmic framework for leveraging side information in imaging via solving black-box multimodal inverse problems in Section~\ref{sec:method}.  Section~\ref{sec:experiments} illustrates the impressive performance of our multimodal diffusion model in reconstructing EBSD images. Finally, we conclude in Section~\ref{sec:conclusion}.

\section{Background \& Related Work}
\label{sec:background}

\subsection{Diffusion models}
Diffusion models \citep{ddpm,song2020score} aim to capture, and sample from, a probability distribution of the data of interest, such as images, and use two Markov processes in \(\mathbb{R}^d\) for this purpose. The forward process gradually adds Gaussian noise to a sample from the data distribution, which is described as
\begin{align}
    x_0 \sim p_{\text{data}}, \quad
    x_t = \sqrt{1 - \beta_t} x_{t-1} + \sqrt{\beta_t} w_t, \quad 1 \leq t \leq T,
\end{align}
where \( \{w_t\}_{1 \leq t \leq T} \) indicates a sequence of independent noise vectors drawn from \( w_t \overset{\text{i.i.d.}}{\sim} \mathcal{N}(0, I_d) \). Here, the parameters \( \{\beta_t \in (0, 1)\} \) represent prescribed learning rate schedules that control the variance of the noise injected in each step. The backward process aims to generate a sample from the data distribution by gradually removing noise from an initial Gaussian sample, in an iterative sampling process described in the following sense: 
\begin{align}
 x_T^{\text{rev}} \rightarrow x_{T-1}^{\text{rev}}    \rightarrow \cdots \rightarrow x_0^{\text{rev}}   .
\end{align}
There exist both deterministic and stochastic samplers for this purpose, which can be interpreted as solving the certain ordinary or stochastic differential equations (ODE/SDEs) in discrete time. Crucially, for both types of samplers, we only need the score functions of the forward process, defined by 
\( s_t^\star(x) = \nabla \log p_{x_t}(x)\). Thanks to a nice property due to Tweedie's formula \citep{efron2011tweedie,vincent2011connection,hyvarinen2007some}, learning the score function at different noise levels is equivalent to Bayes-optimal denoising, i.e., finding the  minimum mean-squared error (MMSE) estimate of \(\epsilon_t\) given \(x_t = x\), 
\begin{align} \label{eq:score_matching}
    s_t^\star(x) := -\frac{1}{1 - \overline{\alpha}_t} \int_{x_0} p_{X_0 \mid X_t}(x_0 \mid x)(x - \sqrt{\overline{\alpha}_t} x_0) \, \mathrm{d}x_0,
\end{align}
where
\begin{align}
    \alpha_t := 1 - \beta_t, \quad \overline{\alpha}_t := \prod_{k=1}^{t} \alpha_k, \quad 1 \leq t \leq T.
\end{align}
In practice, the score functions are estimated via neural network training, a procedure referred to as score matching~\citep{ddpm, song2020score}.

\subsection{Solving inverse problems with diffusion priors}

Due to their expressiveness, diffusion models have been recently promoted as a powerful prior to regularize ill-posed inverse problems. 
A widely adopted paradigm is posterior sampling, where given the observation $y$ in \eqref{eq:forward}, we aim to sample from the posterior distribution
\begin{align}
   p(x \mid y) \propto p^\star(x) p(y \mid x^\star = x) ,
\end{align}
where $p^\star(x)$ is the prior distribution of $x^{\star}$ prescribed by the diffusion model, and $p(y \mid x^\star = x)$ is the likelihood function, typically assumed to be known. Despite a flurry of activities, only a handful posterior sampling algorithms possess provable consistency guarantees, such as \citep{xu2024provably, fpsscm, cardoso2023monte}. In particular, we adopt the recently proposed FPS-SMC algorithm \citep{fpsscm}, which is a provably-consistent particle filtering algorithm tailored for {\em linear} inverse problems.

\section{Leveraging Side Information via Multimodal Black-box Inverse Problems}
\label{sec:method}

Motivated by materials applications, we aim to leverage the availability of side information to accelerate the imaging process, where the side information is given in the form of another low-cost imaging modality described via some black-box forward model without analytical expressions. Let \(x^{\star} \in \mathbb{R}^d  \) be the image of interest from some prior distribution $x^{\star}  \sim p^{\star}(x)$, which is the main imaging modality of interest but expensive to acquire directly. We collect a small number of random measurements of $x^{\star}$, given by
\begin{equation} \label{eq:main}
y_{\mathsf{main}} = \mathcal{P}_{\Omega}(x^{\star}), 
\end{equation}
where $\mathcal{P}_{\Omega}$ is the observation mask preserving the entries in the index set $\Omega$. In addition, we assume we have access to an auxiliary modality in a black-box manner, given by
\begin{equation} \label{eq:auxiliary}
y_{\mathsf{aux}} = f(x^{\star}) + \xi, 
\end{equation}
where  \( f : \mathbb{R}^d \rightarrow \mathbb{R}^m \) can be accessed for functional evaluation without explicit analytical expressions, and $\xi$ is  additional noise. We are interested in developing diffusion-based methods that effectively leverage the side information to improve the quality and speed of imaging. Note that due to the lack of differentiability regarding $f$, it is not possible to directly employ diffusion-based inverse problem solvers targeted at general nonlinear inverse problems such as \citep{xu2024provably}.

\begin{figure}[t]
\centering
\includegraphics[width=0.9\linewidth]{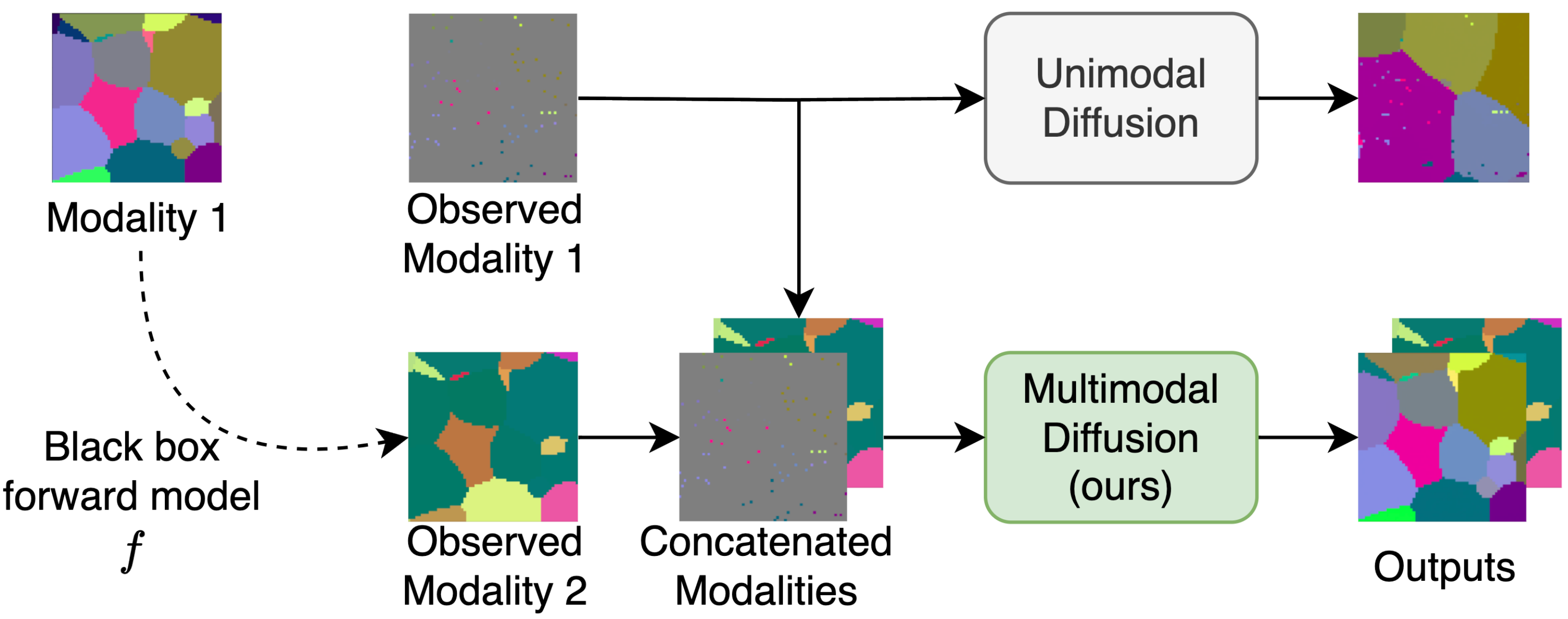}
\vspace{0.05in}
\caption{Method overview. The observed Modality 2 (e.g., PL) is the result of  applying a black box forward model $f$ to Modality 1 (e.g., EBSD), which can only be partially observed. Because of the forward model is unknown, the unimodal model treats this as a inpainting problem and reconstructs Modality 1 poorly. Our multimodal diffusion method concatenates both modalities to reframe the (possibly nonlinear) inverse problem as a linear inpainting problem to produce better results.}
\vspace{-0.05in}
\label{fig:algorithm}
\end{figure}

The above problem can be viewed as a black-box nonlinear inverse problem, which 
we propose a method to reformulate into a linear inpainting inverse problem by training a multimodal diffusion prior, summarized in Figure~\ref{fig:algorithm}. Define the concatenation of the two modalities as
\begin{equation}
X  =\left[ x  \quad f(x ) \right], \qquad x \sim p^{\star} 
\end{equation}
Our approach contains two steps.
\begin{enumerate}
\item {\em Training a multimodal diffusion model.} To train the multimodal diffusion model, we first  
run the black-box forward model $f(\cdot)$ through a training dataset sampled from $p^{\star}$, and merge the two modalities to train a multimodal diffusion model that can sample from the distribution of $X$ via the score matching objective \eqref{eq:score_matching}.

\item {\em Reconstruction via solving an inpainting problem}. Given the observations 
$y_{\mathsf{main}}$ and $y_{\mathsf{aux}}$, we seek to recover $X^{\star} = [ x^{\star} \; f(x^{\star})]$. Note that this can be described as an inpainting forward model,
\begin{align}
   Y = [y_{\mathsf{main}} \; y_{\mathsf{aux}}] = \mathcal{P}_{\bar{\Omega}}  ( X ) + [ 0 \; \xi], 
\end{align}
where $\bar{\Omega}$ corresponds to the augmented observation pattern with the auxiliary modality fully observed.
Consequently, we can invoke linear inverse problem solvers such as FPS-SMC, without the need of accessing the black-box function $f$. 
\end{enumerate}

In this approach, the \textit{score function} of the joint modality implicitly captures the information of the black-box forward model $f$ needed for image reconstruction, bypassing the roadblock of lacking differentiability in the forward process.

\section{Imaging Experiments}
\label{sec:experiments}


We showcase the numerical benefit of incorporating an additional modality into the reconstruction process and solving an inverse problem with a black-box forward model on two modalities defined in different domains via leveraging multimodal diffusion models. In particular, we experiment on the problem of accelerating imaging via electron backscatter diffraction (EBSD) microscopy, an expensive non-Euclidean imaging modality that is slow and cumbersome to collect, from the more easily attainable polarized light (PL) measurements of a material. EBSD images comprise of 3 component vectors describing crystal orientations as Euler angles at each point in space. In contrast, PL microscopy collects c-axis orientation reducible to a 2 component vector for each point in space. Consequently, recovering the full crystal orientation in EBSD data can benefit greatly from the availability of PL data  \citep{jin2018correlation,jin2020c}. There exist works on analyzing and generating EBSD images\citep{dong2023deep,dong2023lightweight,buzzy2024statistically}, but they do not leverage additional modalities. Our goal is to reconstruct the full EBSD image from a small amount of its measurements, and with the availability of PL data, we can dramatically reduce time and cost of data collection.

\subsection{Experimental setup}

We generate synthetic EBSD and PL data using public software, as real data is scarce. First, we create 2 textured 3D EBSD volumes of shape $200 \times 200 \times 400$ using DREAM.3D \citep{groeber2014dream}. Next, for each voxel, we calculate the PL measurements across 36 equal 10$^{\circ}$ rotations using EMsoftOO \citep{emsoft} from the EBSD data; the forward model from EBSD to PL is not available analytically, and treated as a ``black box''. EBSD Euler angles are then converted to cubochoric coordinates \citep{rocsca2014new}, following \citep{dong2023lightweight}. As channels in PL volumes are compressible, we identify all unique voxels in the training data, standardize them, and perform principal component analysis to reduce the dimensionality of all PL data to 3 channels. The EBSD and PL data are then concatenated, so that each voxel has 6 channels (3 for EBSD and 3 for PL). Finally, each channel is transformed to be zero-centered and lie between $-1$ and $1$. For both volumes, the training and validation volumes are of shape $200 \times 200 \times 320$ and $200 \times 200 \times 70$, respectively, with a $200 \times 200 \times 10$ buffer in between them.




\begin{figure}[t]
\centering
\includegraphics[width=0.95\linewidth]{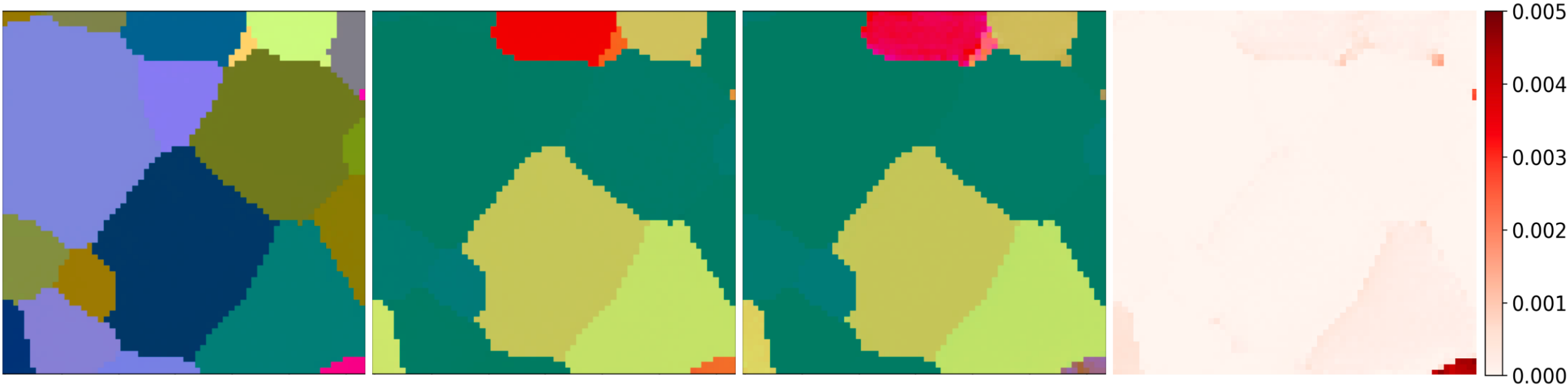} \\
\vspace{0.03in}
\caption{Examples of  generated images from our multimodal diffusion model, where we verify the generated PL data is highly consistent with the PL data by passing through the generated EBSD image through the black-box forward model. From left to right, the images are the generated EBSD, generated PL, PL from applying the forward model on the generated EBSD, and the relative $\ell_2$ consistency error between the PL columns.}
\label{fig:consistency}
\vspace{-0.15in}
\end{figure}

We train a unimodal diffusion model on EBSD data alone and a multimodal diffusion model on EBSD and PL data jointly, using the HuggingFace diffusers library~\citep{von2022diffusers}. 
We use 143 million parameter UNets \citep{ronneberger2015u} for both unimodal and multimodal models with 1 down sampling layer, 1 up sampling layer, and 6 transformer layers for both unimodal and multimodal diffusion models. The parameter difference is negligible between the two and is only present in the first layer due to a different input dimension. Random $64 \times 64$ slices are sampled from the validation volumes to create the validation set. For training data, we do the same 2D sampling from volumes at each gradient step. We train both models for 30000 gradient steps with batches of 128 samples, taking around 12 hours on a single Nvidia L40 GPU. 

To confirm the multimodal training quality and how it captures the forward model, Figure~\ref{fig:consistency} shows that for the unconditionally generated EBSD and PL images, when we pass the generated EBSD through the true forward model, the resulting PL data is very close to the PL data generated by the multimodal model. This experiment confirms that the trained multimodal diffusion model successfully captures the relationship between modalities and learns the black-box forward model implicitly. 


In the following experiments, we observe a small percentage of EBSD entries for both the unimodal and multimodal models uniformly at random. For the multimodal model, we fully observe the PL data, potentially with heavy Gaussian noise to examine the robustness of our approach. We chose FPS-SMC \citep{fpsscm} as the reconstruction method, which is tailored to linear inverse problems and has theoretical guarantees for convergence.
Even though the model being trained on cubochoric coordinates, we evaluate the reconstruction error in the original non-Euclidean domain using Euler angle disorientation \citep{larsen2017improved,ebsdtorch}.

\subsection{Performance}

\begin{figure}[!t]
\centering
\includegraphics[width=0.7\linewidth]{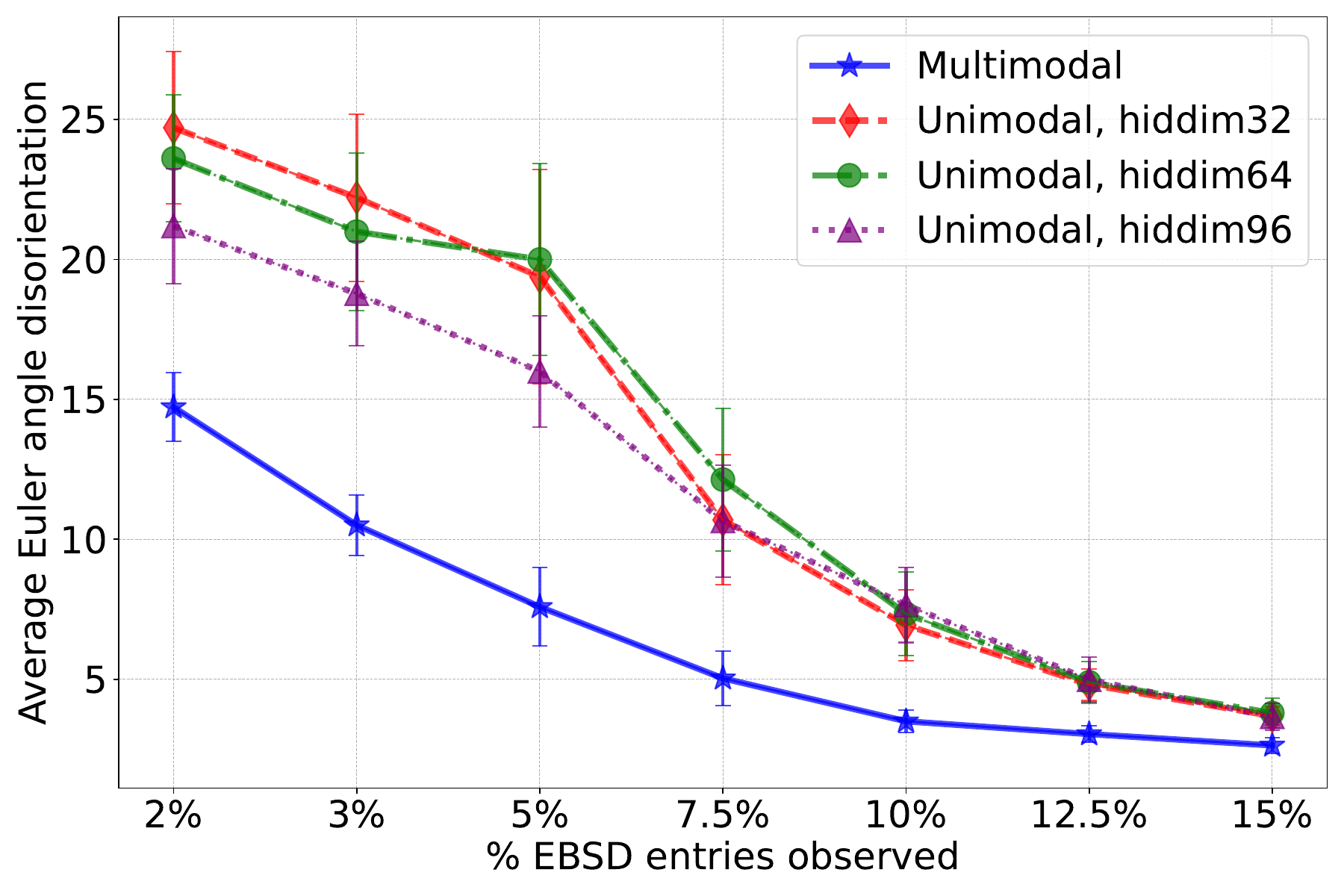}
\caption{Comparison of the unimodal and multimodal performance (measured by Euler angle disorientation) with respect to the fraction of observed EBSD entries for both models. We use three varying sizes with increasing hidden dimension for the unimodal UNet with the largest being the same size as the multimodal UNet. Lower disorientation is better.}
\label{fig:res}
\vspace{-0.1in}
\end{figure}

\begin{figure}[!h]
\centering
\includegraphics[width=0.7\linewidth]{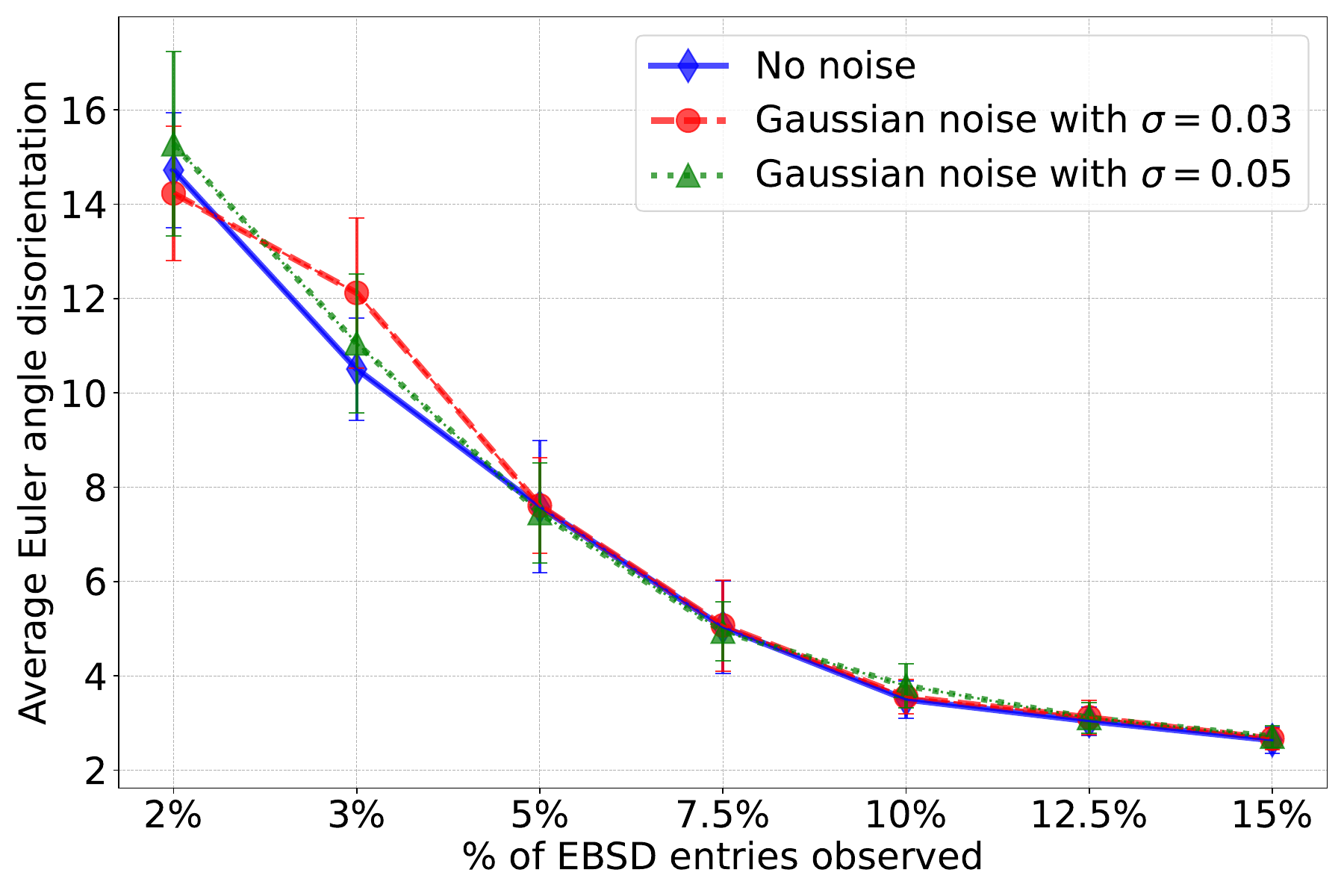}
\caption{Performance comparison of our multimodal diffusion model for different amounts of PL measurement noise injected while observing 2\% noiseless EBSD. Lower disorientation is better. Our multimodal model is robust to noise in PL observations, as performance across varying noise levels are similar to the noiseless case.
}
\label{fig:res2}
\vspace{-0.1in}
\end{figure}

\begin{figure}[h]
\centering
\includegraphics[width=0.7\linewidth]{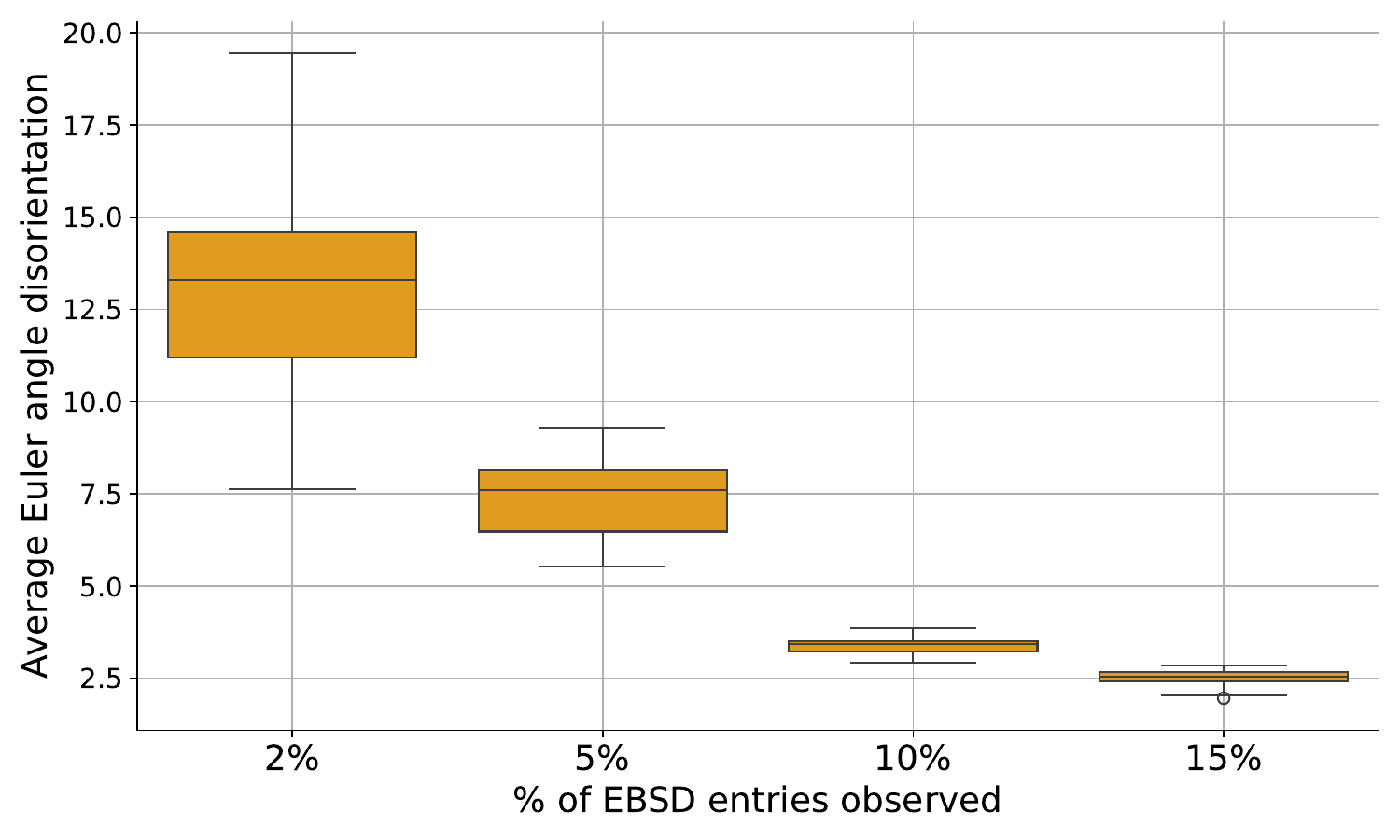}
\caption{The uncertainty quantification of the reconstruction error across 20 generated EBSD images from the same observations at different observation levels of EBSD entries, where the full noiseless PL image is observed for all cases.
}
\label{fig:variancePlot}
\end{figure}

\begin{figure}[!h]
\centering
\hspace{0.1in}
\includegraphics[width=0.16\linewidth]{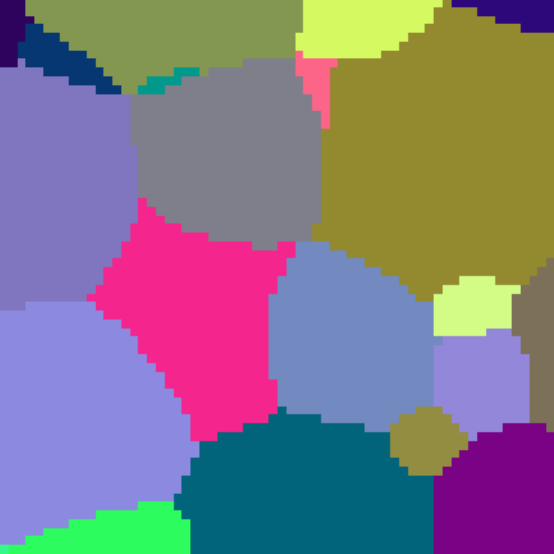}
\hspace{0.4in}
\includegraphics[width=0.16\linewidth]{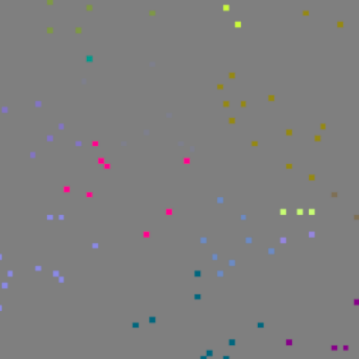}\\
\quad Ground truth EBSD \quad Observed EBSD \\
\vspace{0.04in}
\includegraphics[width=0.9\linewidth]{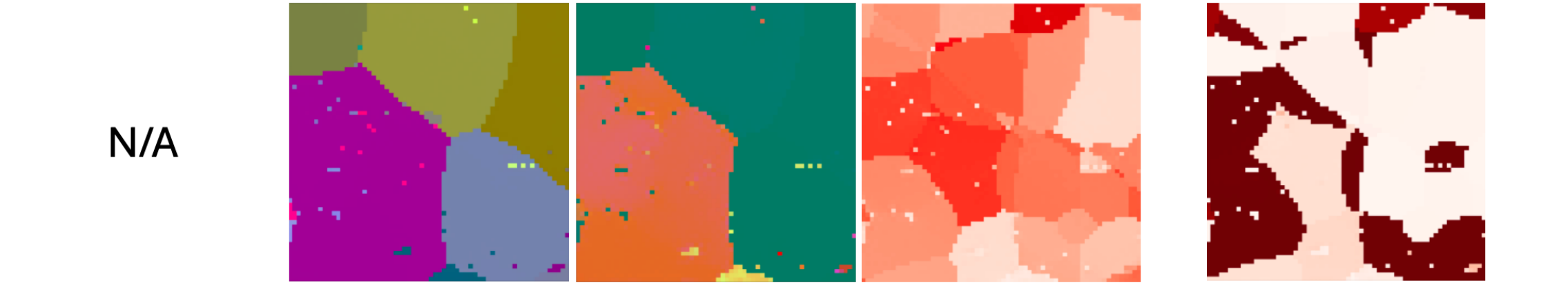}\\
Unimodal, without PL observations \\
\vspace{0.04in}
\includegraphics[width=0.9\linewidth]{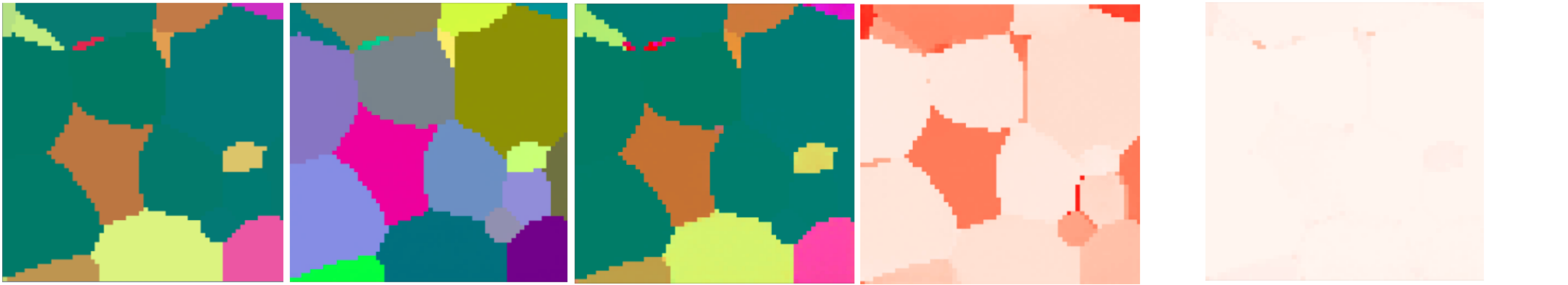}\\
Multimodal, noiseless PL observations \\
\vspace{0.04in}
\includegraphics[width=0.9\linewidth]{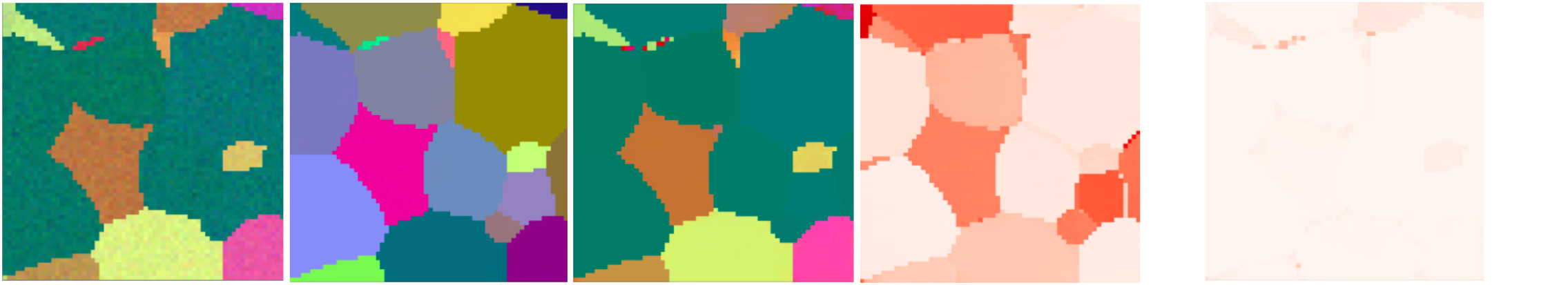}\\
Multimodal, noisy PL observations ($\sigma = 0.05$) \\
\vspace{0.04in}
\includegraphics[width=0.9\linewidth]{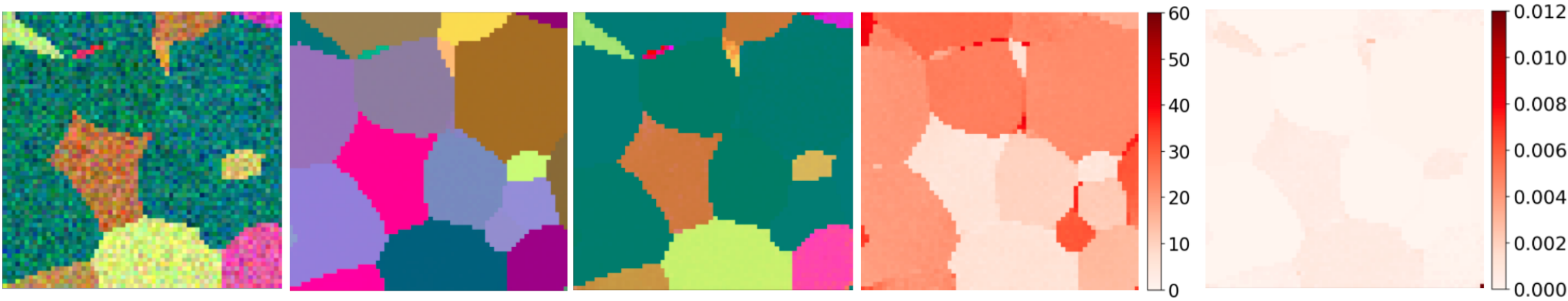}\\
Multimodal, noisy PL observations ($\sigma = 0.2$) \\
\vspace{0.04in}
\caption{Example EBSD reconstructions by the unimodal model and multimodal model (without and with PL observations). From left to right, the columns are the observed PL, the generated EBSD, the PL from feeding the generated EBSD into the forward model, the disorientation error, and the relative $\ell_2$ consistency error between the PL columns. Darker shades of red in the right-most two columns indicate higher error. Note that due to the geometry of EBSD data, visual closeness in color for the generated EBSD does not imply low disorientation error and vice versa.}
\label{fig:example_outputs}
\vspace{-0.1in}
\end{figure}

In Figure~\ref{fig:res}, we observe that the multimodal model outperforms the unimodal model by a large margin in terms of the reconstruction error, measured by the average Euler angle disorientation. For fair comparison, the number of parameters is held the same for the largest unimodal model and the multimodal model, as well as the amount of training data fed into both models. Both the unimodal and multimodal algorithms take approximately 75 seconds to reconstruct a single image, so incorporating the additional modality adds no noticeable computational cost at inference time.

Next, we investigate the influence of noise on the image reconstruction with the multimodal model, by adding Gaussian noise on the PL data. As seen in Figure~\ref{fig:res2}, increasing the noise level does not have a big influence on the model performance especially as we collect more EBSD entries. Qualitative examples in Figure~\ref{fig:example_outputs} show that, given only \(2\%\) of observed EBSD entries, the reconstruction provided by the multimodal model with the additional PL modality is significantly better than the reconstruction provided based on only the EBSD entries using the unimodal model. The boundaries are correctly detected, and the disorientation is high only for only a few regions with our model. On the other hand, the unimodal model fails to correctly reconstruct the boundaries. Moreover, using the reconstructed EBSD data from the multimodal model, we obtained their corresponding PL data by running the forward model, and there was little deviation from the  observed PL measurements. In addition, our multimodal model is robust to the PL measurement noise. Even with Gaussian noise with standard deviation $\sigma=0.05$, reconstruction quality is similar to the noiseless case. For a very large Gaussian noise with standard deviation \(\sigma=0.2\), it still gets most of the boundaries and EBSD values with graceful performance degeneration.

To quantify the uncertainty level of the reconstruction, we sample multiple EBSD images from the same noiseless observations and evaluate the reconstruction error for each generation. From Figure~\ref{fig:variancePlot}, we can see that with more observed EBSD entries, the variance of the reconstruction error also significantly decreases for the same observations, which indicates that the variability of the reconstruction can be used to indicate its quality.

\section{Conclusion}
\label{sec:conclusion}

We demonstrate the effectiveness of multimodal diffusion models to reframe inverse problems with black-box forward models into a linear inpainting task. By learning a joint prior on multiple modalities, we can eliminate the need to have access to the forward model at inference time and deploy any inverse problem solver for inpainting. Even though our multimodal model is trained nearly identically to the unimodal model, it observes superior reconstruction quality which is validated on a materials microscopy case study with non-Euclidean data and a black-box forward model, allowing us to skip much of the data collection for the expensive microscopy modality. Additionally, our multimodal model is robust to heavy noise and produces results highly consistent with the observations, capturing the forward model and further demonstrating the efficacy of incorporating information from other modalities. For future work, we aim to explore training with geometric constraints, black-box inverse problems in other scientific settings, reconstruction with 3D data instead of 2D data, and further reconstruction acceleration using faster samplers~\citep{liaccelerating, lu2022dpm}.

\newpage
\section*{Acknowledgements}

The work of T. Efimov, H. Dong and Y. Chi is supported in part by the Air Force D3OM2S Center of Excellence under FA8650-19-2-5209, by ONR under N00014-19-1-2404, by NSF under ECCS-2126634, and by the Carnegie Mellon University Manufacturing Futures Initiative, made possible by the Richard King Mellon Foundation. The work of H. Dong is also supported by the Wei Shen and Xuehong Zhang Presidential Fellowship at Carnegie Mellon University.

\bibliographystyle{apalike}
\bibliography{reference-diffusion}

\end{document}